\documentclass[a4paper, 10pt, conference]{IEEEtran}
\IEEEoverridecommandlockouts
\usepackage{url}
\usepackage{amsmath, amsthm, amssymb}
\usepackage{abraces}
\usepackage{balance}
\usepackage{graphicx}
\usepackage{subfigure}
\usepackage{color}
\usepackage{indentfirst}
\usepackage{yhmath}
\usepackage{multirow}
\usepackage{stfloats}
\usepackage{array} 
\usepackage{threeparttable}
\usepackage{algorithm}
\usepackage{algorithmic}
\usepackage[square, comma, sort&compress, numbers]{natbib}
\newcolumntype{L}[1]{>{\raggedright\let\newline\\\arraybackslash\hspace{0pt}}m{#1}} 
\newcolumntype{C}[1]{>{\centering\let\newline\\\arraybackslash\hspace{0pt}}m{#1}} 
\newcolumntype{R}[1]{>{\raggedleft\let\newline\\\arraybackslash\hspace{0pt}}m{#1}}

\title{\LARGE \bf Model predictive approach to integrated path planning and tracking for autonomous vehicles}
\author{Chao Huang$^{1}$, Boyuan Li$^{2}$  and Masako Kishida$^{1}$
\thanks{$^{1}$ Chao Huang and Masako Kishida are with the National Institute of Informatics, Tokyo, 101-8430, Japan. {\tt\small chao$\_$huang@nii.ac.jp}, {\tt\small kishida@nii.ac.jp}}
\thanks{$^{2}$ Boyuan Li is with State Key Laboratory of Advanced Design and Manufacture for Vehicle Body, Hunan University, Changsha, 410082, China. {\tt\small boyuanli@hnu.edu.cn}}
}

\begin{document}
\maketitle
\thispagestyle{empty}
\pagestyle{empty}
\begin{abstract}
In the path planning problem of autonomous application, the existing studies separately consider the path planning and trajectory tracking control of the autonomous vehicle and few of them have integrated the trajectory planning and trajectory control together. To fill in this research gap, this study proposes an integrated trajectory planning and trajectory control method. This paper also studies the collision avoidance problem of autonomous by considering static and dynamic obstacles. Simulation results have been presented to show the effectiveness of the proposed control method.
\end{abstract}
\section{Introduction}
Autonomous vehicles have a promising future of making transportation time effortless and enabling the drivers to partake in other activities and therefore change everyday life across the world. They have the potential to dramatically reduce crashes which are caused by driver's faults including aggressive driving, over-compensation, inexperience, slow reaction times, inattention, and various other driver shortcomings. In addition, the autonomous vehicle can provide mobility to the elderly and disabled, increase road capacity, save fuel, and lower harmful emissions \cite{fagnant2014preparing}.

Experiments have been conducted on autonomous vehicles since 1920s: promising trials took place in the 1950s and work has proceeded since then. Many large global companies such as Google, Tesla, BMW and Lexus are investing fund and resources in the research of autonomous vehicles. Google’s co-founder Sergey Brin at Recode Code Conference in Palos Verdes, California described the concept of autonomous vehicles \cite{griffiths2004shared}. The prototype of autonomous vehicles relies on built-in sensors and a software system to safely manoeuvre the vehicle without manual controls. 

One of several key problems to be solved before autonomous vehicle becoming a reality is to determine a collision-free trajectory of the autonomous vehicles based on the sensor information of sensing and mapping module. In general, the autonomous vehicle control is divided into path planning (trajectory generation) and trajectory tracking. The path planning problem is to find a desired path for the vehicle that starts at the initial configuration and reaches the goal region while satisfying given global and local constraints. Effective path planning algorithms are what make autonomous driving genuinely feasible, safe, and fast \cite{paden2016survey}. Meanwhile, the path tracking problem is to design control laws for the individual vehicle's actuator to achieve the selected trajectory. 

The real-time path planning, however, particularly in the presence of obstacles, remains very challenging \cite{shen2017integrated}. There exist many different approaches for solving the path-planning problem. The graph-search approaches such as Dijkstra algorithm \cite{kala2013multi,bohren2009little}, $A^\star$ algorithm \cite{ziegler2008navigating}, and State Lattice algorithm \cite{rufli2009application}, construct graphical discretization of the vehicle’s state space and search for a shortest path using graph search methods. The incremental tree-based approaches incrementally construct a tree of reachable states from the initial state of the vehicle and then select the best branch of such a tree. The sampling based approaches have been proven to be an effective framework that is suitable for a large class of problems in domains. Existing methods in  robotics such as Probabilistic  Roadmap  Method  (PRM)  \cite{kavraki1994probabilistic}  and  the Rapidly-exploring  Random  Tree  (RRT)  \cite{lavalle2001randomized} are extensively tested for automated vehicles.  However, these methods are computationally expensive. The  interpolating  curve  planners  implement different techniques for path smoothing and curve generation by considering feasibility, comfort, vehicle dynamics, and other parameters in  the  automated  driving  field \cite{horst2006trajectory, brezak2014real}.  

In the path tracking controller, many of the studies assume the desired vehicle path is already known or the desired path has been planned by the off-line trajectory planner. Specifically, the autonomous vehicle is planned to follow the given trajectory, which is assumed to be collision-free and can be achieved by the vehicle \cite{griffiths2004shared}. In addition, tracking controller designed based on linearized or oversimplified models can ignore important nonlinear dynamics that play a major role when the vehicle is operated close to the limits of its handling capability \cite{paden2016survey}.

Most of these studies separately consider the trajectory planning and trajectory control of the autonomous vehicle and few of them have integrated the trajectory planning and trajectory control together expect a few \cite{shen2017integrated}.  We design a control method that combines the path planning and tracking by using a nonlinear vehicle Model Predictive (MP) algorithm in this paper. This integrated MP algorithm simultaneously optimises the front wheel angle and brake torque directly and predicts the local path in real time. The additional tracking controller is no longer required and the computational efficiency can be improved by this integrated structure. MP algorithm has been extensively applied in real-time path planning and tracking control and has been proven to generate the desired safety path. Most of these studies only use the single point mass model and assume the vehicle has a constant longitudinal speed. This assumption can hardly describe the actual vehicle dynamics and the vehicle dynamics performance may be seriously compromised when following the planned trajectory. In this paper, the MP algorithm is based on 2-degree-of-freedom (2-DOF) non-linear bicycle model. First, we calculate a Dubins path by considering the acceleration limits of the car. Compared with road centre line, the Dubins path is proven to be shortest and flexible. Then, in the optimisation cost function of the proposed MP algorithm, in addition to minimise the distance from the road centre line and maximize the distance from the road boundary and obstacles, we also consider the smoothness and comfort of the path by minimizing the yaw acceleration of the vehicle throughout the whole manoeuvre. The obstacles on the road can be classified as the static obstacles and dynamic obstacles and the proposed integrated method can control the vehicle to avoid the obstacles simultaneously in a fast and efficient manner.

The remainder of the paper is structured as follows. Section \ref{setion2} provides models of the dynamics of vehicle. The driving manoeuvre and geometric path are designed in Section \ref{section3}  and the the MP-based integrated trajectory planning and control
algorithm are introduced in Section \ref{section4}. Simulation results are provided in Section \ref{section5}. Finally, Section \ref{section6} concludes this paper. 

\section{Vehicle dynamics model}\label{setion2}
The use of autonomous vehicles brings with a number of advantages compared to human operated vehicles \cite{pepy2006path}, and autonomous vehicles have already been developed and produced for production, construction, and urban environments. 

The dynamics of the vehicle is modeled by a bicycle model as shown in Fig. 1:

\begin{equation}
\begin{aligned}
\label{equ1}
\dot{v}_x&=rv_y-\frac{2}{m}(F_{cf}\sin{\delta_f}-T_r/R_w)\\
\dot{v}_y&=-rv_x+\frac{2}{m}(F_{cf}\cos{\delta_f}+F_{cr})\\
\dot{r}&=\frac{2}{I_z}(l_fF_{cf}-l_rF_{cr})\\
\dot{X}&=v_x\cos{\psi}-v_y\sin{\psi}\\
\dot{Y}&=v_x\sin{\psi}+v_y\cos{\psi}
\end{aligned}
\end{equation}
where $x$ and $y$ are the coordinates of the  centre of mass (CM) in an inertial frame. $v_x=\dot{x}$ and $v_y=\dot{y}$ are the longitudinal and lateral speeds of CM in an inertial frame. $\Psi$ is the inertial heading angle and $r = \dot{\Psi}$ is the yaw rate. $m$ and $I_z$ denote the vehicle’s mass and yaw inertia, respectively. $l_f$ and $l_r$ represent the distance from CM of the vehicle to the front and rear axles, respectively. $\dot{X}$ and $\dot{Y}$ denote the longitudinal and lateral speeds of CM in the body frame $(X, Y)$, respectively. $F_{cf}$ and $F_{cr}$ denote the lateral tyre forces at the front and rear wheels, respectively, in the inertial frame aligned with the wheels. It is assumed that 
these lateral tyre forces are expressed by
\begin{equation}
\label{equ2}
F_{ci}=-C_{\alpha_i} \alpha_i,\ \ i \in f, r
\end{equation}
where $C_{\alpha_i}$ are the front and rear tyre cornering stiffness, and $\alpha_i$ are the front and rear slip angles given by
\begin{equation}
\begin{aligned}
\alpha_f &= \dfrac{v_y+l_fr}{v_x} -\delta_f\\
\alpha_r &= \dfrac{v_y-l_rr}{v_x} 
\end{aligned}
\end{equation}
with a small angle assumption.



\begin{figure}[t!]
	\begin{center}	{\includegraphics[width=0.38\textwidth]{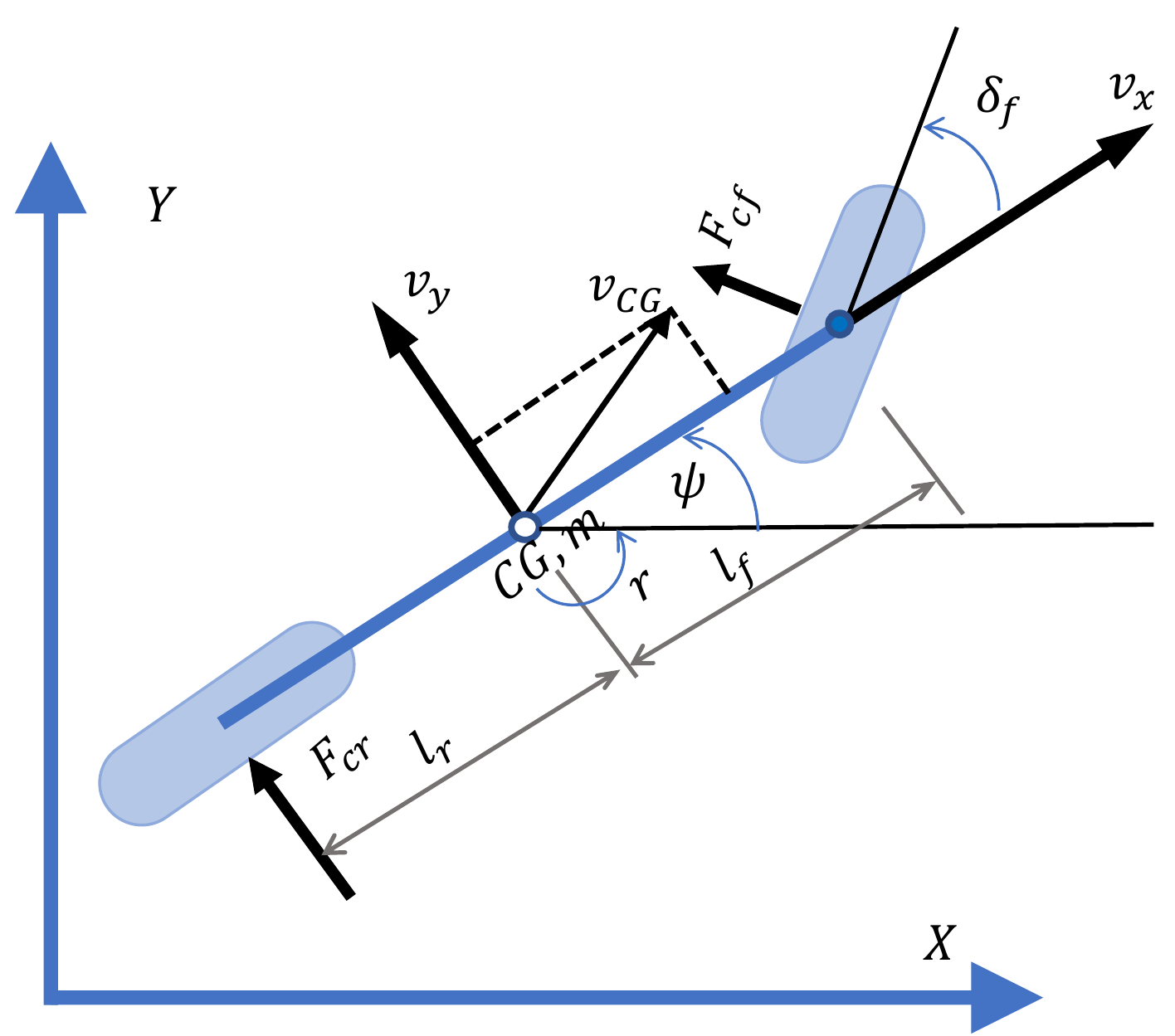}}
		\caption{Bicycle model of vehicle}
		\label{fig:bicycle}
	\end{center}
\end{figure}

\newcounter{TempEqCnt}
\setcounter{TempEqCnt}{\value{equation}}
\setcounter{equation}{4}
\begin{figure*}
\begin{equation}
\begin{aligned}
\label{5}
&\min_{\delta_f(k)\cdots \delta_f(k+N_p-1), T_r(k) \cdots T_r(k+N_p-1)}{J}=\sum_{i=1}^{N_p} \Biggl\{
a_1[(\hat{x}_a(k+i)-X_d(k+i))^2+(\hat{y}_a(k+i)-Y_d(k+i))^2]  \\
 &+b_1\left[\frac{1}{(\hat{x}_a(k+i)-X_u(k+i))^2+(\hat{y}_a(k+i)-Y_u(k+i))^2}\right]^2\\ 
&\left.+b_2\left[\frac{1}{(\hat{x}_a(k+i)-X_l(k+i))^2+(\hat{y}_a(k+i)-Y_l(k+i))^2}\right]^2+b_3\left | \frac{d\hat{r}(k+i)}{dt}\Big |_{t=\varepsilon} \right | ^2 \right\}\\
\end{aligned}
\end{equation}
s.t.\\
\begin{equation*}
\begin{array}{c}
-\delta_{max}\leq \delta_f (k+i) \leq \delta_{max}\\
-T_{bmax} \leq  T_r (k+i)  \leq T_{dmax}
\end{array}\  \ \ \ \text{for all}\ \ i=1 \cdots N_p
\end{equation*} 
\end{figure*}

\section{Driving manoeuvre}

In this paper, we design a single lane change manoeuvre. The single lane change manoeuvre is a common test for vehicle handling as it represents an essential collision avoidance manoeuvre, especially on the motorway.  The road information including road width, road boundary and the obstacles information including vehicles's speed are assumed to be known.

\subsection{Dubins path}\label{section3}
One of the most popular path planning methods found in robotics is the geometric method. Dubins (1957) showed that, for a particle that does not reverse, the shortest paths consist of circular arcs and straight line segments. This section describes the construction of such paths suitable for the target vehicle to perform specified lateral obstacle avoidance manoeuvre.

It is known that the vehicle is capable of a maximum acceleration of $\mu g\ [m/s^2]$ \cite{bevan2008development} and this will result in a circular path with radius $R$
\begin{equation}
\label{equ4}
R=\dot{X}^2/(\mu g), 
\end{equation}
where $\dot{X}$ is the longitudinal speed of the vehicle, $\mu$ is the friction coefficient and $g$ is the gravity constant.

\begin{figure}[h]
	\begin{center}	{\includegraphics[width=0.42\textwidth]{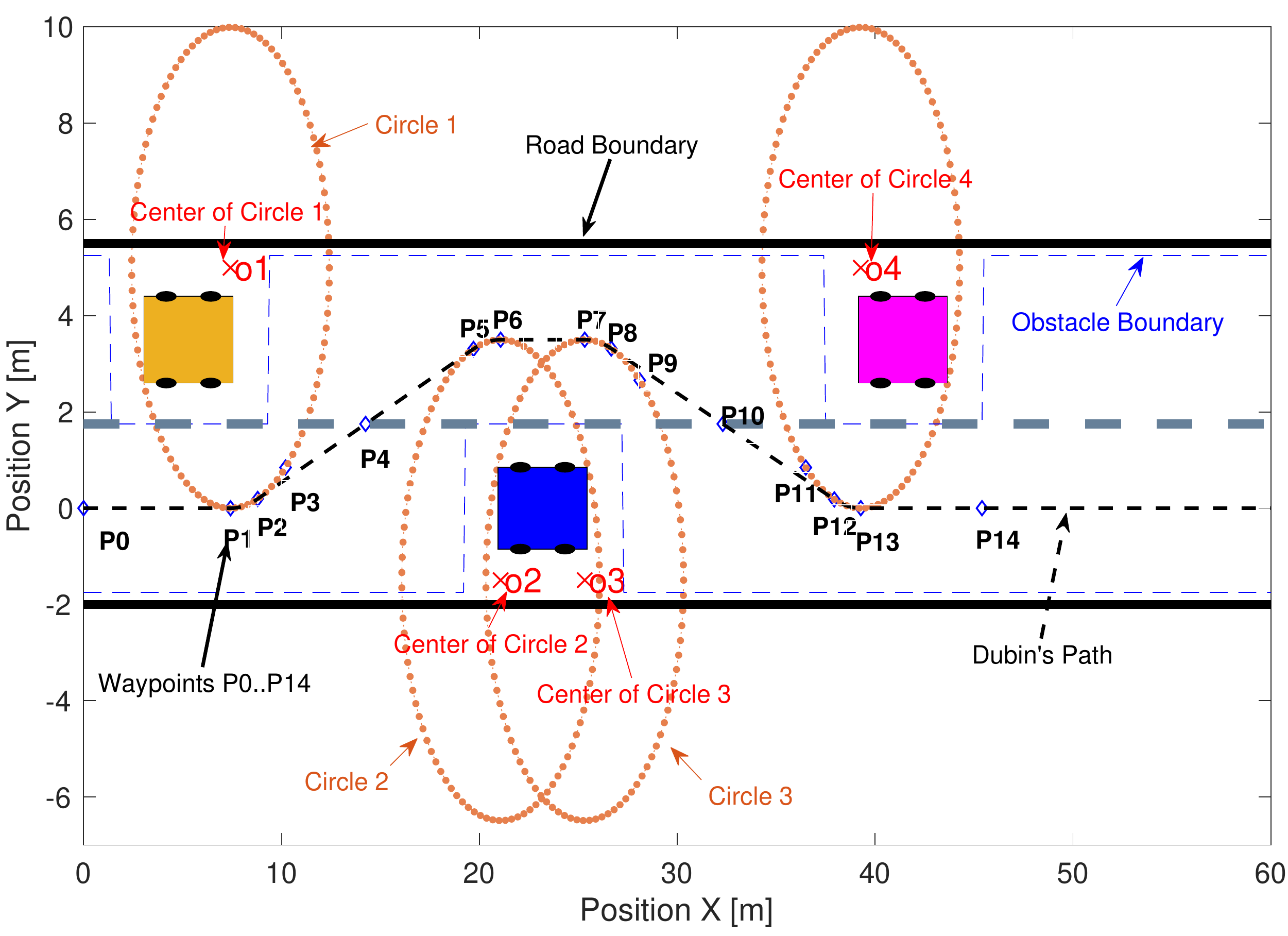}}
		\caption{Geometric construction of reference trajectory}
		\label{fig:geo}
	\end{center}
\end{figure}

Fig. \ref{fig:geo} shows the construction of a Dubins path consisting of straight lines and circular arcs for a single lane change. The road width is $3.5m$ and its boundaries are represented by black solid lines. There are three obstacles (yellow vehicle, pink vehicle and blue vehicle) and it is assumed that based on the sensor information, all the obstacles can be detected by the ego vehicle and the current position and velocity of each obstacle can be accurately estimated.  Therefore, the obstacle boundary is known at each time step and represented by blue dash line. It is noticed that the obstacle boundary has considered the safety gap between the ego vehicle and obstacle vehicles. The grey dash line presents the road  centre line.  The initial position of the ego vehicle is $P_0=(0,0)$.

By knowing the sensor information of road and obstacles and $R$ in (\ref{equ4}), we define some waypoints like $P_1, P_2, \cdots P_{14}$ which are the keys to describe the Dubins path. Specifically, the path is designed so that the ego vehicle drives passing the initial position $P_0$ and follows the centre of the lower lane until reaching $P_1$, which is the beginning of the maximum acceleration turn to the upper lane. Continuing the turn through $P_2$, the point of the closest approach to the boundary, the vehicle reaches $P_3$, from where it follows a straight path from $P_4$ to $P_5$. The maximum acceleration turn to the lower lane, through $P_6$ and $P_7$, brings the vehicle to the centre of the adjacent lane. From $P_8$ to $P_{11}$, the vehicle starts the second turn and eventually ends to the centre of the original lane ($P_{12}$ to $P_{14}$). After defining all the waypoints, the whole path containing straight lines and arcs can be completely described. It is noted that under the premise of obtaining the information of obstacles, the path can feasibly avoid static and dynamic obstacles in the lane change manoeuvre. More details can be found in \cite{bevan2008development}.

\subsection{MP optimisation}\label{section4}
However, the transition between arc and line segments entails discontinuous changes in lateral acceleration, making it impossible for real autonomous vehicle to track exactly. In this paper, MP algorithm is used to obtain control inputs $\delta_f$ and $T_r$ to minimize the error.

Based on the 2 DOF tyre model, the cost function of MP optimisation algorithm at time $k$ can be presented in (\ref{5}).
In  (\ref{5}), $N_p$ is the prediction horizon, $X_d$ and $Y_d$ are the longitudinal and lateral positions of the Dubins path which are generated in Section \ref{section3}. $X_u$ and $Y_u$ are the longitudinal and lateral positions of road upper boundary and $X_l$ and $Y_l$ are the longitudinal and lateral positions of road lower boundary. The $\hat{x}_a$ and $\hat{y}_a$  are the predicted trajectory optimised by this MP algorithm. $a_1$ is scaling factor of attractive potential and $b_1$, $b_2$ and $b_3$ are scaling factors of repulsive potential. Here, $a_1$ is related to the term of minimizing the distance between the vehicle position and Dubins path. $b_1$ and $b_2$ are related to the terms of maximizing the distance between the vehicle position and road boundaries. $b_3$ is related to the term of minimizing yaw acceleration throughout the manoeuvre. The values of these scaling factors can be adjusted according to the priority of different scenarios. The optimisation variables are control inputs, front wheel steering angle $\delta_f(k)\cdots \delta_f(k+N_p-1)$ and rear wheel traction or brake torque $T_r(k) \cdots T_r(k+N_p-1)$. 

The constraints are placed on the steering angle and driving or braking torque. In general, the front wheel angle of a conventional vehicle is constrained by the mechanical linkages between each wheel. According to \cite{karnopp2004vehicle}, the constraints on the front wheel angle for a traditional front wheel steering vehicle is between $+35^{\circ}$ and $-35^{\circ}$. For four-wheel-independent-steering vehicles \cite{li2014novel}, the front wheel angle can achieve a rotation of $\pm 90^{\circ}$ while driving. In this paper, the front wheel angles of Steer-by-Wire (SbW) systems can lie only within $-45^{\circ}$ and $45^{\circ}$. Any degree outside of this range is invalid and the vehicle will not be able to turn to that amount. The maximum driving torque of the real wheel of vehicle model $T_{dmax}$ is $200Nm$ and the total braking torque $T_{bmax}$ is $160Nm$.

The $\hat{x}_a$ and $\hat{y}_a$ are the future vehicle status which can be predicted and calculated from the vehicle velocity in the previous time step based on the following discrete dynamics model:
\setcounter{equation}{\value{TempEqCnt}}
\setcounter{equation}{5}
\begin{equation}
\begin{aligned}
\hat{x}_a(k+i)&=x_g(k+i-1)\\
&+\hat{v}_{xg}(k+i)\left(t(k+i)-t(k+i-1) \right),\\
\hat{y}_a(k+i)&=y_g(k+i-1)\\
&+\hat{v}_{yg}(k+i)\left(t(k+i)-t(k+i-1)\right)\\
&\text{with}\ i=1\cdots N_p
\end{aligned}
\end{equation}
where $x_g(k)$ and $y_g(k)$ are feedback values of vehicle longitudinal and lateral position in the global coordinate system in the previous time step. Here, $t(k+i-1)$ is the time of the current time step and $t(k+i)$ is the time of the next time step. $\hat{v}_{xg}(k+i)$ and $\hat{v}_{yg}(k+i)$ are predicted vehicle longitudinal and lateral velocities in the global coordinate system in the current time step, which can be presented as follows:
\begin{equation}
\begin{aligned}
\hat{v}_{xg}(k+i)&=\hat{v}_x(k+i)\cos(\hat{\psi}(k+i))-\hat{v}_y(k+i)\sin(\hat{\psi}(k+i)),\\
\hat{v}_{yg}(k+i)&=\hat{v}_x(k+i)\sin(\hat{\psi}(k+i))+\hat{v}_y(k+i)\cos(\hat{\psi}(k+i))\\
&\text{with}\ i=1\cdots N_p
\end{aligned}
\end{equation}
where $\hat{v}_x(k+i)$ and $\hat{v}_y(k+i)$ are predicted longitudinal and lateral velocities in the body-fixed coordinate system and $\hat{\psi}(k+i)$ is the predicted vehicle yaw angle. According to (\ref{equ1}), $\hat{v}_x(k+i)$, $\hat{v}_y(k+i)$ and $\hat{r}(k+i)$  can be predicted in (8).

\setcounter{TempEqCnt}{\value{equation}}
\setcounter{equation}{7}
\begin{figure*}
\begin{equation}
\begin{array}{l}
\label{equ7}
\hat{v}_x(k+i)=v_x(k+i-1)+\left (v_y(k+i-1)r(k+i-1)-\frac{2 \hat{F}_{cf}(k+i-1) \sin \delta_f (k+i-1)-T_r(k+i-1)/R_w}{m} \right ) (t(k+i)-t(k+i-1)),\\
\hat{v}_y(k+i)=v_y(k+i-1)+\left ( -v_x(k+i-1)r(k+i-1)+\frac{2 \hat{F}_{cf}(k+i-1) \cos \delta_f (k+i-1)+\hat{F}_{cr}(k+i-1)}{m}  \right ) (t(k+i)-t(k+i-1)),\\
\hat{r}(k+i)=r(k+i-1)+\left ( \frac{2 l_f \hat{F}_{cf}(k+i-1)-l_r \hat{F}_{cr}(k+i-1)}{m} \right ) \left ( t(k+i)-t(k+i-1) \right )\\
\hat{\psi}(k+i)=\psi(k+i-1)+r(k+i-1)(t(k+i)-t(k+i-1)), \text{with}\ i=1\cdots N_p
\end{array}
\end{equation}
\end{figure*} 

In (8), $v_x(k)$, $v_y(k)$ and $r(k)$ are feedback values from actual vehicle in the current time step. $\hat{F}_{cf}(k+i-1)$ and $\hat{F}_{cr}(k+i-1)$ are predicted tyre front wheel side force, rear wheel side force in the previous time step. On the basis of (\ref{equ2}), $\hat{F}_{cf}(k+i-1)$ and $\hat{F}_{cr}(k+i-1)$ can be predicted as follows:
\setcounter{equation}{\value{TempEqCnt}}
\setcounter{equation}{8}
\begin{equation}
\label{equ7}
\begin{array}{l}
\hat{F}_{cf}(k+i-1)=-C_{\alpha_f}\left(\frac{v_y(k+i-1)+l_fr(k+i-1)}{v_x(k+i-1)}-\delta_f(k+i-1) \right),\\
\hat{F}_{cr}(k+i-1)=-C_{\alpha_r}\left(\frac{v_y(k+i-1)-l_rr(k+i-1)}{v_x(k+i-1)} \right)\\
\text{with}\ i=1\cdots N_p
\end{array}
\end{equation}

In this section, the vehicle trajectory in real time can be predicted and optimised by (5)–(9) with the real-time feedback values from actual vehicle. The optimisation variables of steering angle and driving torque can be determined and input into the actual vehicle model. This method is summarized in Algorithm 1.

\begin{algorithm}
\caption{The proposed algorithm}
\label{alg:A}
\begin{algorithmic}[1]
\STATE{Set prediction horizion $N_p$}
\FOR{$t=t_{initial}$ to $t_{finial}$}
\STATE {Generate Dubins path based on Section \ref{section3};} 
\STATE {Solve optimisation problem (\ref{5}); }
\STATE {Go back to 2 with $t=k+1$}.
\ENDFOR
\end{algorithmic}
\end{algorithm}

\section{Simulation results}\label{section5}

In this section, two sets of simulations are carried out by the software of MATLAB/Simulink to verify the advantages of the proposed integrated trajectory planning and tracking controller based on MP algorithm and bicycle nonlinear model. The vehicle parameters are listed in TABLE 1.

\begin{table}[htb]
	\begin{center}
		\caption{Parameters of vehicle dynamics for simulation }\label{table:valuevehicledynamics} 
		\begin{tabular}{|C{1.6cm}|C{1.5cm}|}
			\hline
			\textbf{Parameter} &\textbf{Value} \\\hline
			$l_f\ (m)$&1.2 \\\hline
			$l_r\ (m)$&1.05\\\hline
			$m (kg)$& 2000\\\hline
			$I_z (kg\cdot m^2)$& 1300\\\hline
			$\mu$&0.5\\\hline
			$C_{\alpha f}\ (N/rad)$&12000\\\hline
			$C_{\alpha r}\ (N/rad)$&12000\\\hline
		\end{tabular}
	\end{center}
\end{table}

\begin{figure}
	\begin{center}	{\includegraphics[width=0.40\textwidth]{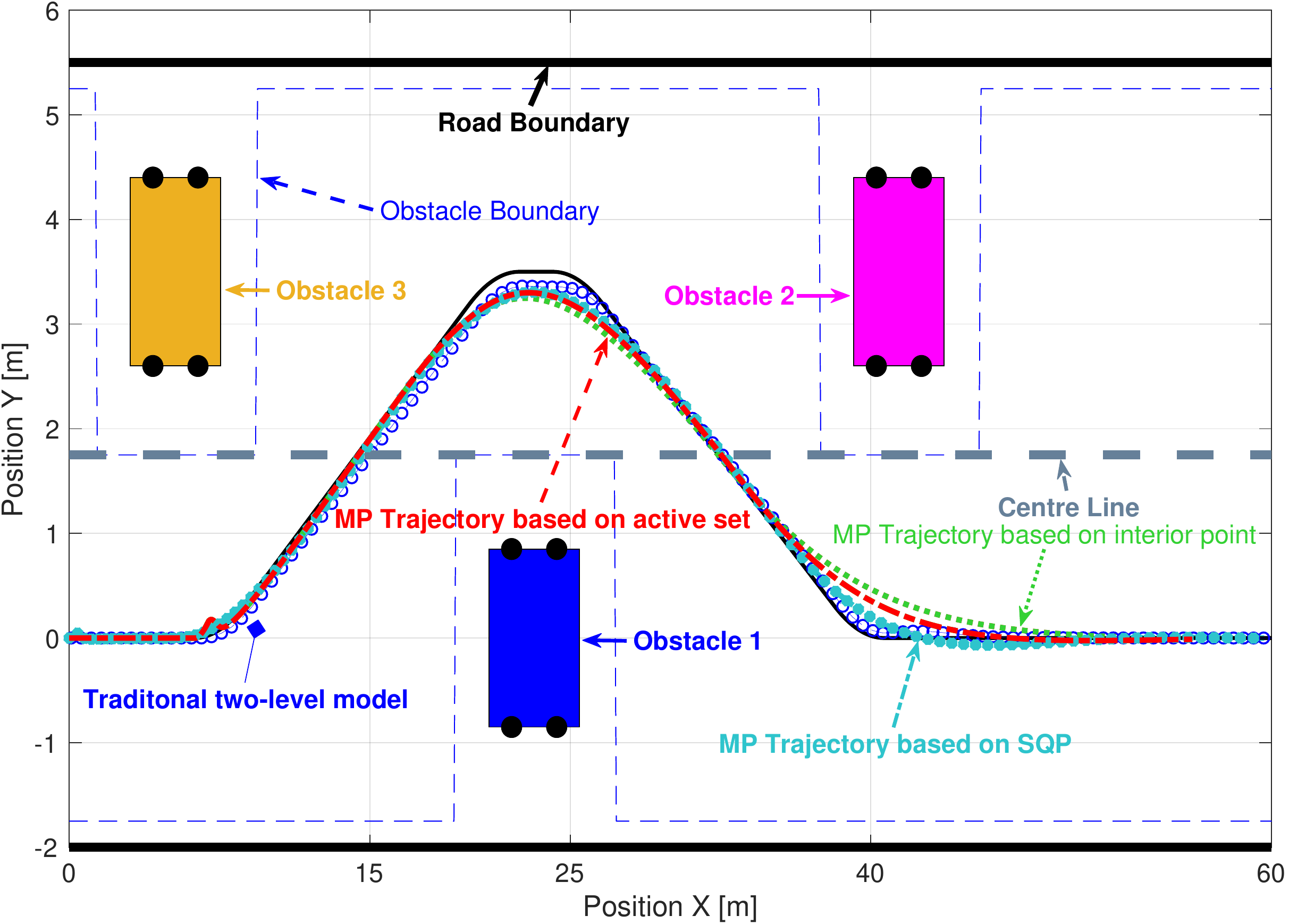}}
		\caption{Vehicle desired and actual trajectory with different algorithms}
		\label{fig:difff}
	\end{center}
\end{figure}

\begin{figure}
	\begin{center}	{\includegraphics[width=0.40\textwidth]{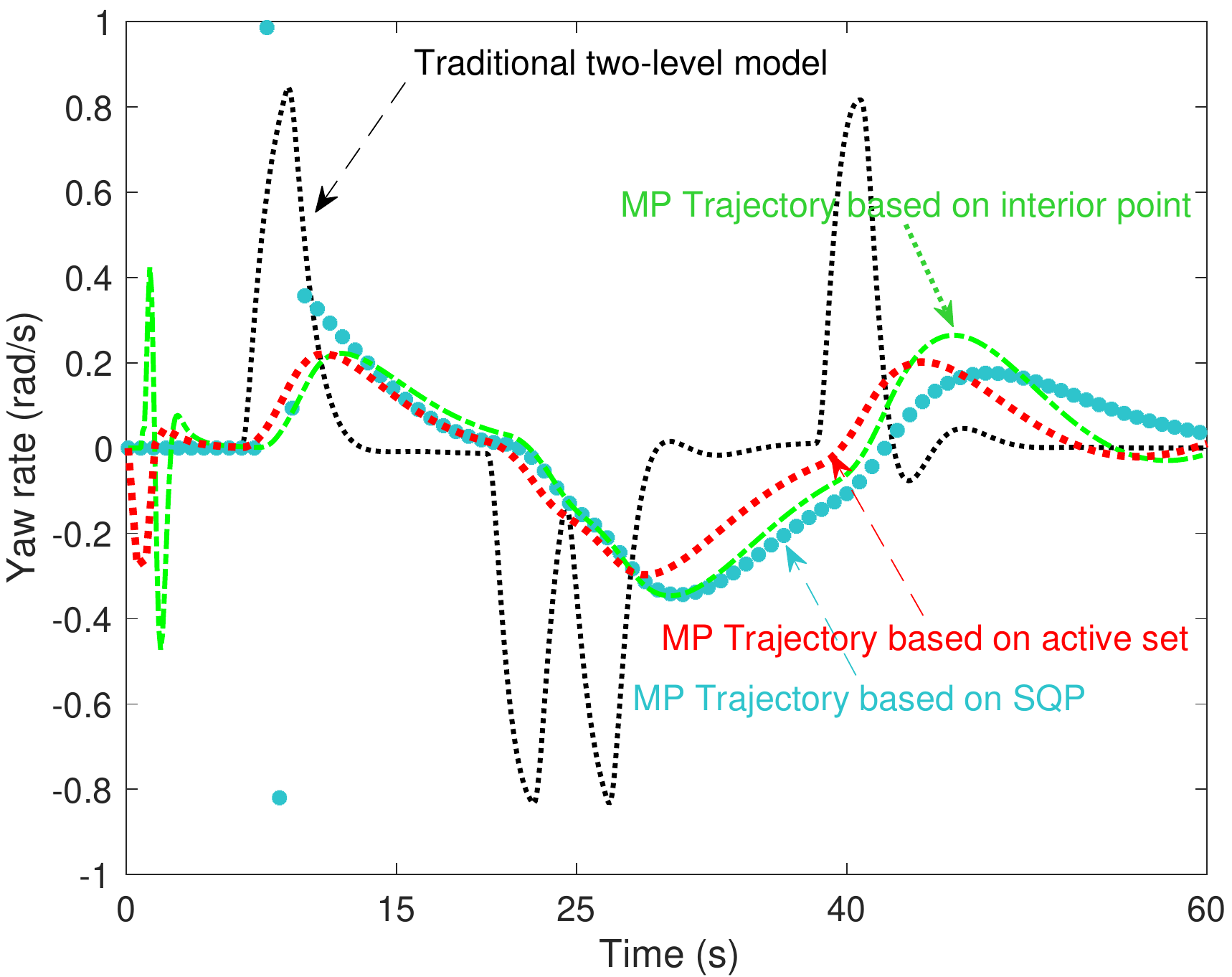}}
		\caption{Yaw rate with different algorithms}
		\label{fig:difffyaw}
	\end{center}
\end{figure}

The standard Matlab function $\textit{fmincon}$ with receding control strategy is implemented to solve (\ref{5}) with the sampling time $0.1s$. The prediction horizon is $N_p=3$. 

Three algorithms for $\textit{fmincon}$: interior point, sequential quadratic programming (SQP) and active set algorithms are compared with the traditional two-level method in Fig. \ref{fig:difff} and Fig. \ref{fig:difffyaw}. It is assumed that the initial velocity of the autonomous vehicle is $10m/s$ and all the obstacles are static. under the assumption that the obstacles are static. Compared with traditional two-level method, the MP trajectories can provide a smooth and flexible path. We select the active set algorithm (red dash line) in the following simulations.

\begin{figure}
	\begin{center}	{\includegraphics[width=0.42\textwidth]{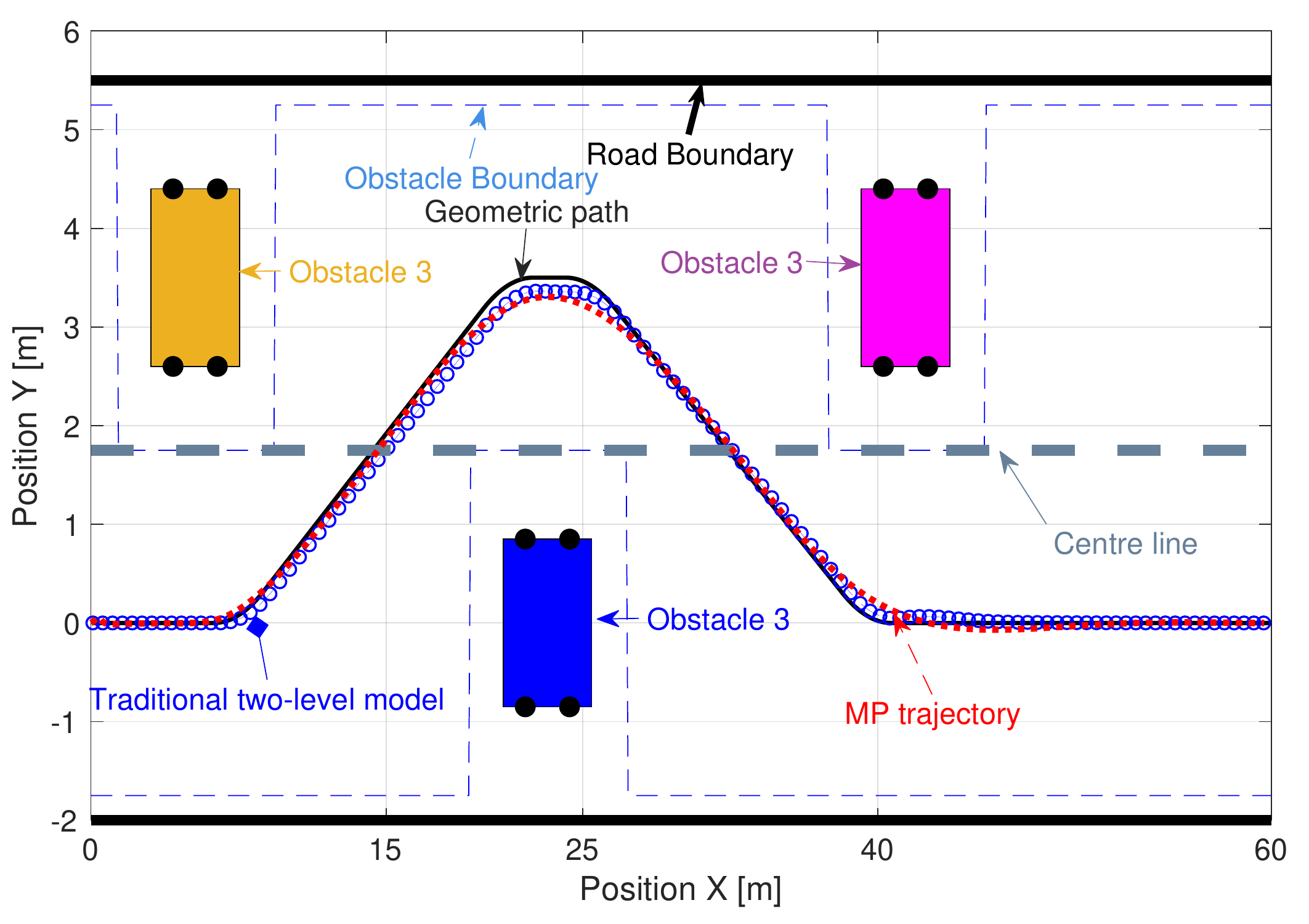}}
		\caption{Vehicle desired and actual trajectory in the first set of simulations}
		\label{fig:static}
	\end{center}
\end{figure}

\begin{figure}
	\begin{center}	{\includegraphics[width=0.42\textwidth]{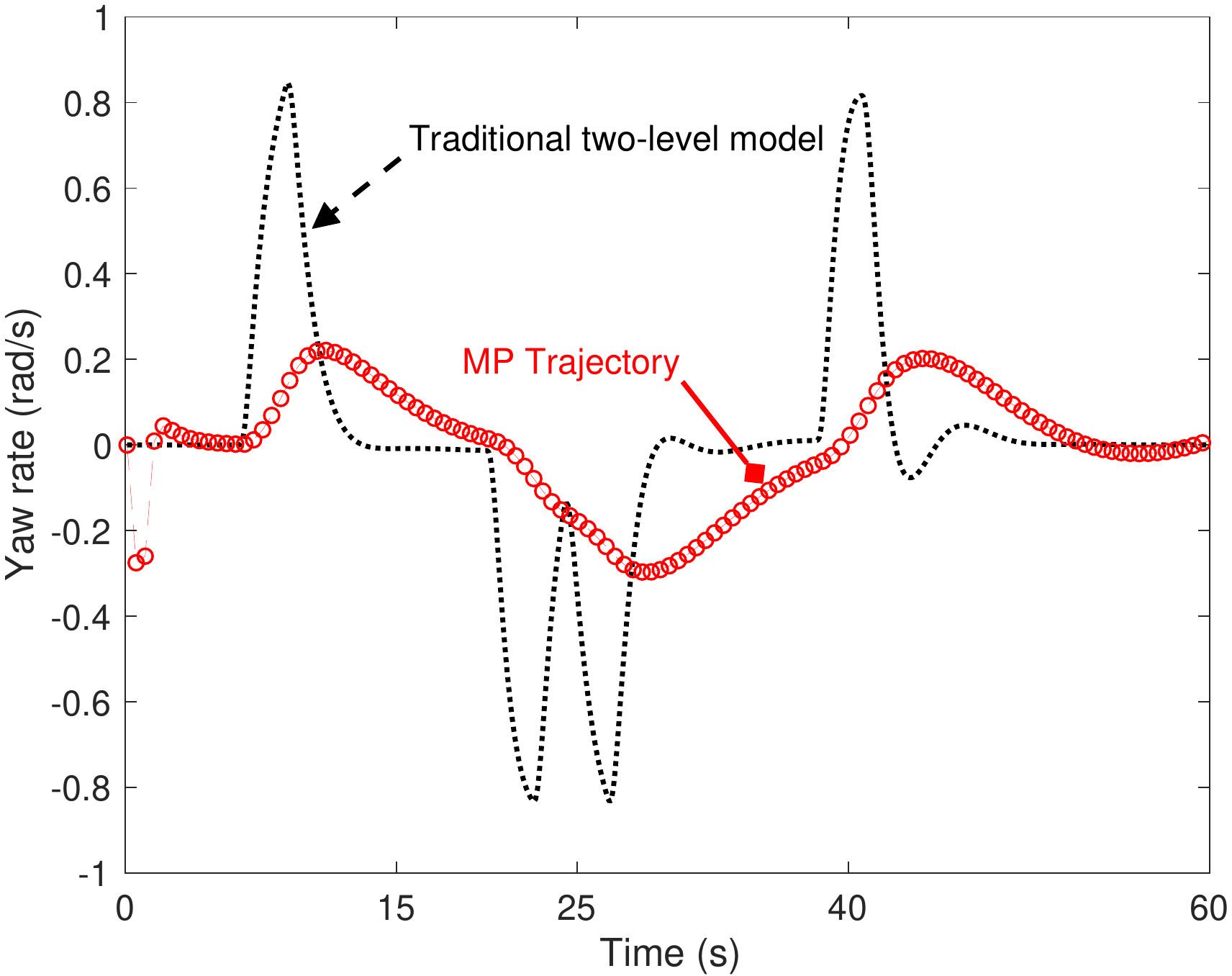}}
		\caption{Yaw rate in the first set of simulations}
		\label{fig:yaw}
	\end{center}
\end{figure}

Fig. \ref{fig:static} and Fig. \ref{fig:yaw} show the first set of simulation where the obstacles are static. The initial velocity of the autonomous vehicle is 10 $m/s$. The actual trajectory is controlled by the proposed MP algorithm based on the  2 DOF tyre model. Furthermore, the traditional two-level method is compared and shown in Fig. \ref{fig:static}.  In the traditional two-level method, the desired trajectory is assumed to be planned and known already and the two-level trajectory tracking controller is implemented to achieve the desired path. Fig. \ref{fig:static} also shows that the trajectories of all these two method can accurately follow the Dubins path and the proposed MP method can provide a smoother lane change path. Fig. \ref{fig:yaw} shows the vehicle yaw rate response of these methods and the proposed MP method shows more stable yaw rate response and less yaw angle change rate compared with traditional two-level method.  In contrast,  the yaw rate response of traditional two-level method is oscillating abruptly and the reason behind is that the desired path is pre-defined and the trajectory cannot be smoothly optimised in real time.

\begin{figure}
	\begin{center}	{\includegraphics[width=0.42\textwidth]{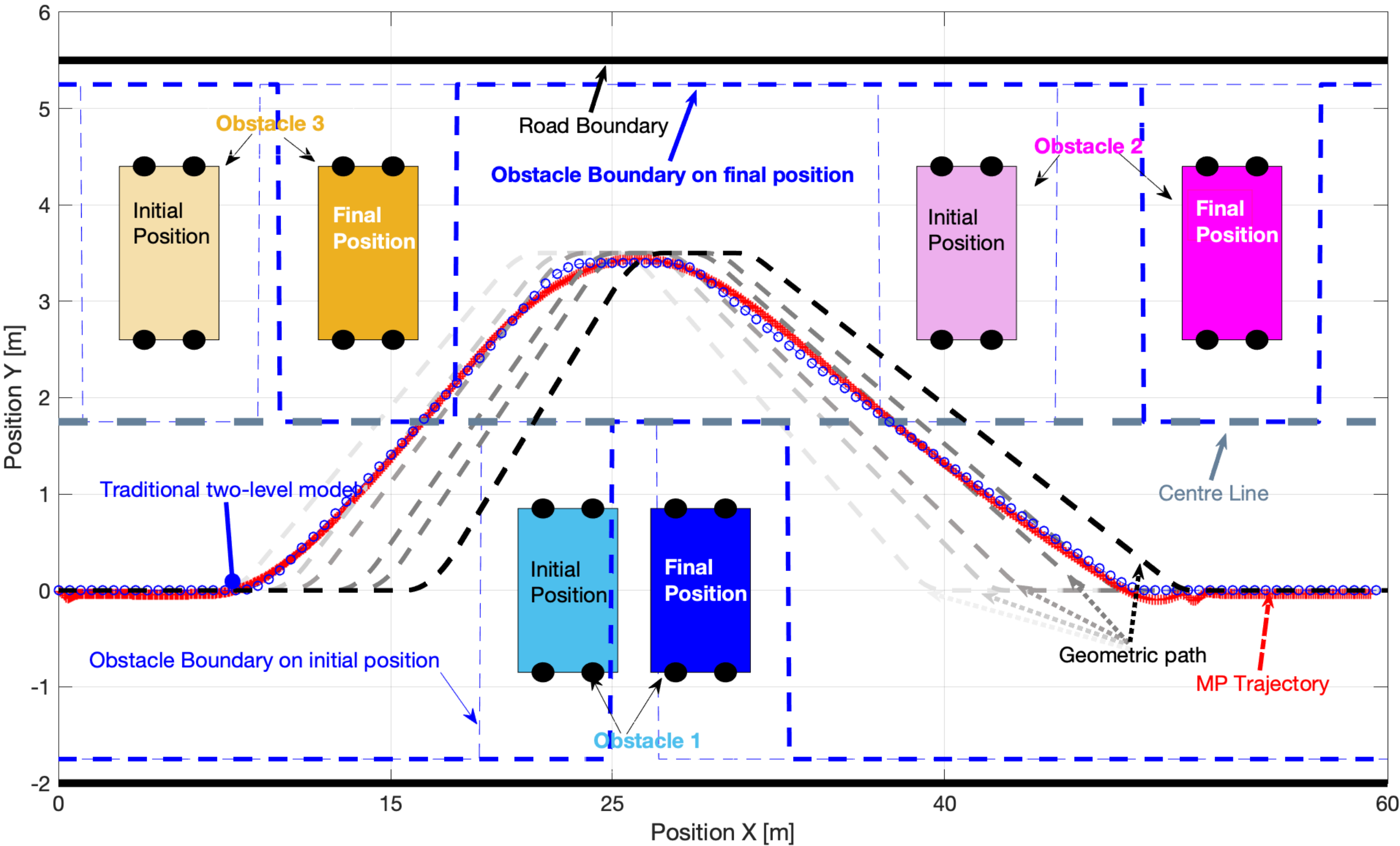}}
		\caption{Vehicle desired and actual trajectory in the second set of simulations}
		\label{fig:dynamic}
	\end{center}
\end{figure}

\begin{figure}
	\begin{center}	{\includegraphics[width=0.42\textwidth]{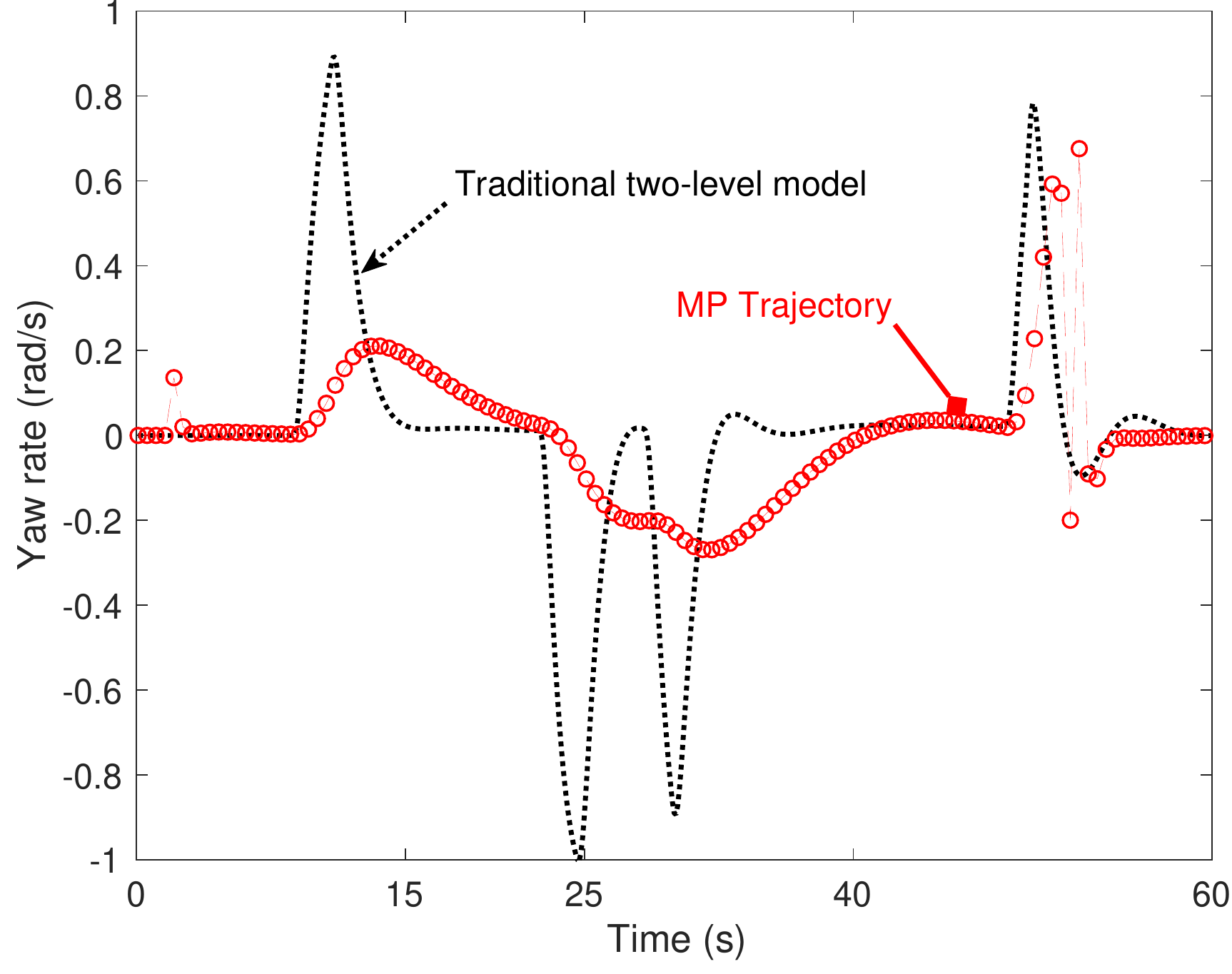}}
		\caption{Yaw rate in the second set of simulations}
		\label{fig:yaw2}
	\end{center}
\end{figure}

Fig. \ref{fig:dynamic} and Fig. \ref{fig:yaw2} show the second set of simulation, the autonomous vehicle is still implementing the lane change task with the same boundary condition in Fig. 2, but, the obstacles are dynamic. The initial velocity of all these three vehicles are 10 $m/s$. The obstacle 1 will accelerate and drive at 10.5 $m/s$ and obstacle 2 will accelerate and drive at 12 $m/s$ and obstacle 3 will accelerate and drive at 11.5 $m/s$.  All the parameters of the ego vehicle are the same as that in the first set of simulation. It is noted that instead of following the pre-defined path in the first set of simulation, the traditional two-level method applies the MP method to generate the obstacle avoidance trajectory. Fig. \ref{fig:dynamic} shows the actual vehicle trajectory of the proposed method and traditional methods and Fig. \ref{fig:yaw2} shows the yaw rate response in the second set of simulation with dynamic obstacles. It is indicated that the proposed MP algorithm can successfully generate the avoidance trajectory and also generate a smooth lane change path.

\section{CONCLUSIONS}\label{section6}

This paper proposes a real-time control method that integrates path planning and tracking based on the MP algorithm for autonomous vehicles. In the future, we are interested in extending the approach for multiple autonomous vehicles.
\section*{ACKNOWLEDGMENT}
The authors are supported by ERATO HASUO Metamathematics for Systems Design Project (No.{JPMJER1603}), JST.

\bibliography{mpcfinal}{}
\bibliographystyle{IEEEtran}
\end{document}